\documentclass[10pt, a4paper]{article}

\usepackage[final]{lrec-coling2024} % this is the new style
\usepackage{times}
\usepackage{latexsym}
\usepackage{graphicx}
\usepackage{bm}
\usepackage{multirow} 
\usepackage{amsmath}
\usepackage{amssymb}
\usepackage{amsfonts}
\usepackage{booktabs}
\usepackage{footnote}

\title{\normalfont{\textbf{CARE: Co-Attention Network for Joint Entity and Relation Extraction}}}

\name{Wenjun Kong and Yamei Xia$^{\ast}$ \thanks{$^{\ast}$Corresponding author.}}

\address{School of Computer Science, Beijing University of Posts and Telecommunications \\
         \{kongwenjun, ymxia\}@bupt.edu.cn\\}

\abstract{
Joint entity and relation extraction is the fundamental task of information extraction, consisting of two subtasks: named entity recognition and relation extraction. However, most existing joint extraction methods suffer from issues of feature confusion or inadequate interaction between the two subtasks. Addressing these challenges, in this work, we propose a \textbf{C}o-\textbf{A}ttention network for joint entity and \textbf{R}elation \textbf{E}xtraction (CARE). Our approach includes adopting a parallel encoding strategy to learn separate representations for each subtask, aiming to avoid feature overlap or confusion. At the core of our approach is the co-attention module that captures two-way interaction between the two subtasks, allowing the model to leverage entity information for relation prediction and vice versa, thus promoting mutual enhancement. Through extensive experiments on three benchmark datasets for joint entity and relation extraction (NYT, WebNLG, and SciERC), we demonstrate that our proposed model outperforms existing baseline models. Our code will be available at \url{https://github.com/kwj0x7f/CARE}.
 \\ \newline  }

\begin{document}

\maketitleabstract

\section{Introduction}

\begin{figure}
    \centering
    \includegraphics[width=1.0\linewidth]{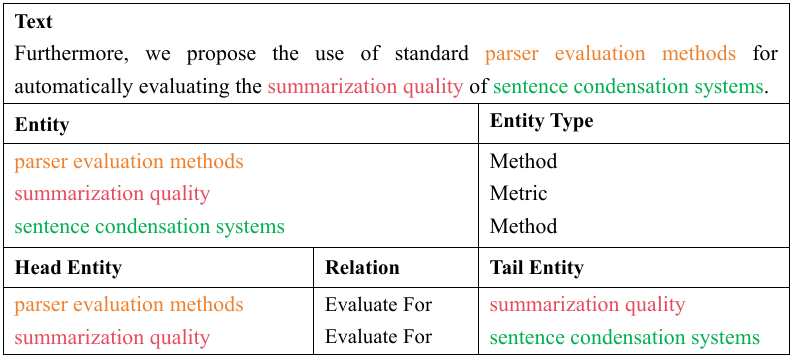}
    \caption{An example from the SciERC dataset \citep{luan2018multi}. Entities are distinguished by being assigned various colors.}
    \label{fig:running-example}
\end{figure}

Named entity recognition (NER) and relation extraction (RE) are fundamental for various NLP applications like information retrieval, sentiment analysis and question answering. These tasks empower computers to comprehend the underlying meanings within text, thereby enabling more sophisticated comprehension and manipulation of linguistic information. Figure~\ref{fig:running-example} shows an example from the SciERC dataset \citep{luan2018multi}, exemplifying the task of identifying named entities and entity-relation triplets. 

Traditionally, NER and RE were approached as separate stages in a processing pipeline \citep{zhong2021frustratingly}. In this pipeline approach, the text would first undergo NER, where entities like names, locations, and dates are identified and categorized. Next, the output of NER would be passed to the relation extraction stage. The pipeline approach struggles to capture complex interactions between subtasks. This limitation arises due to the isolated nature of the pipeline stages, where errors or ambiguities from one stage can propagate to subsequent stages, leading to cascading errors.

End-to-end or joint modeling approaches \citep{eberts2020span,wang2020two,wei2020novel,yan2021partition,wang2021unire} aim to mitigate these limitations. By learning the interdependencies between NER and RE tasks in a joint model, these approaches can better capture the intricate connections between entities and relations. 

Unfortunately, existing joint approaches still exhibit certain limitations. Some methods \citep{bekoulis2018joint,eberts2020span} encounter feature confusion due to the utilization of shared representations across both subtasks. In addition, most existing joint models \citep{bekoulis2018joint,eberts2020span,wei2020novel} fail to effectively model the interaction between NER and RE. Considering that entity information often carries significant relational indicators, and conversely, relation information contains valuable cues about entities, this oversight results in suboptimal performance.

To overcome the above limitations, effective approaches are needed that disentangle features, enhance representation learning, and foster a more harmonious interaction between the two interconnected tasks. To this end, we address the following two main challenges:

\begin{figure*}
    \centering
    \includegraphics[width=1.0\linewidth]{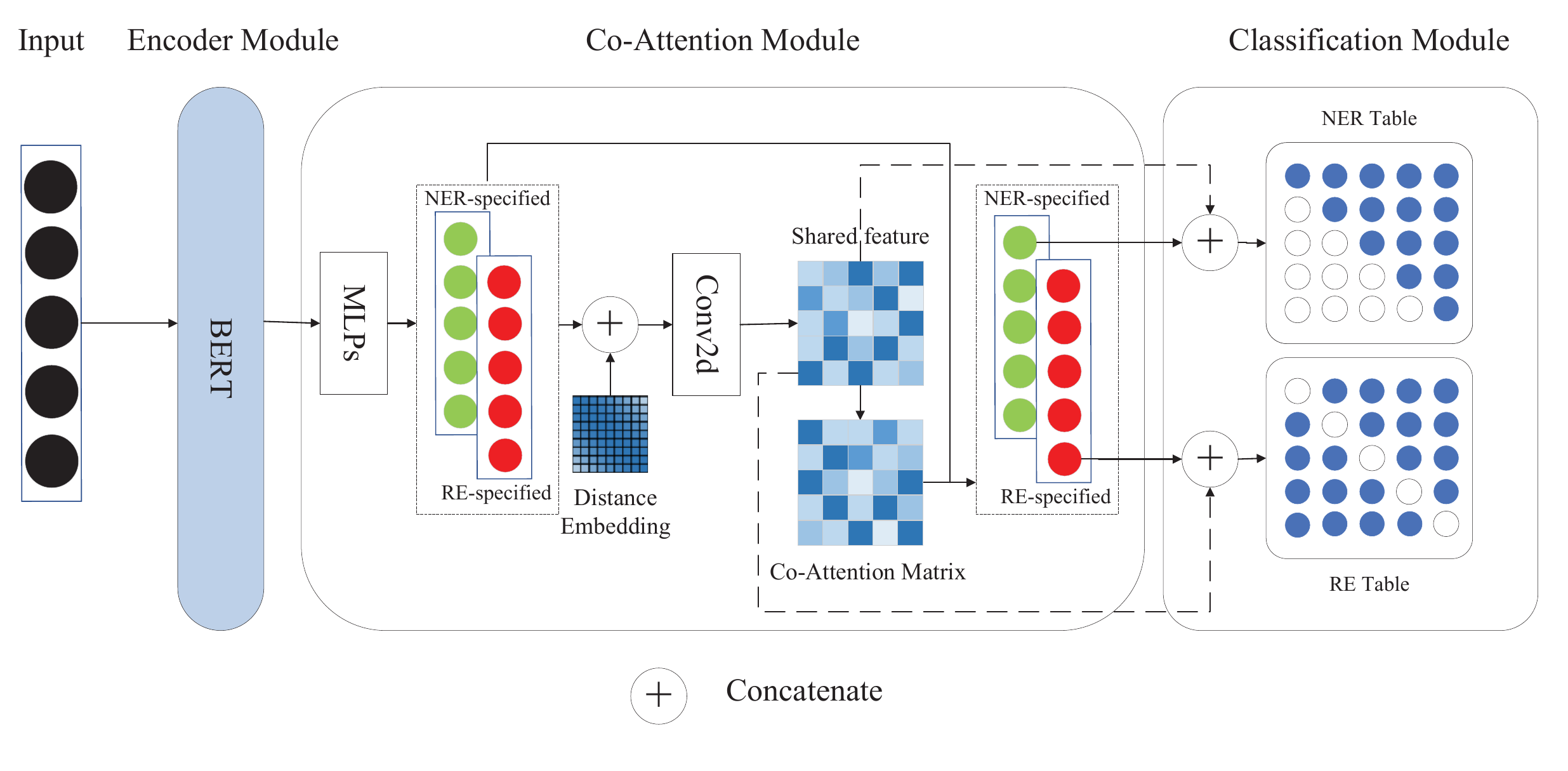}
    \caption{The overall framework of CARE, which consists of three modules: encoder module, co-attention module, and classification module. NER-specified and RE-specified representations are colored as green and red respectively.}
    \label{fig:model}
\end{figure*}

Firstly, how to alleviate feature confusion within the joint model when learning features across the two subtasks? Our approach involves employing task-specific multi-layer perceptrons, enabling the acquisition of task-specified representations for each subtask independently, thereby avoiding feature overlap or confusion.

Secondly, how to effectively model the intricate interaction between the two subtasks within a joint framework? We propose to employ a co-attention mechanism to model two-way interaction between NER and RE. Therefore, entity information can be harnessed to predict relations, and vice versa, enabling mutual enhancement between the two subtasks.

In summary, our main contributions are as follows:
\begin{itemize}

    \item We propose CARE, a \textbf{C}o-\textbf{A}ttention network for joint entity and \textbf{R}elation \textbf{E}xtraction, which can effectively exploit interactions between subtasks.
    
    \item Extensive experiments on three benchmark datasets show that our model can achieve superior performance compared with existing methods, and ablation studies demonstrate the effectiveness of our method.

\end{itemize}

\section{Problem Definition}
\paragraph{Joint Entity and Relation Extraction.} The input of this problem is a sentence $X$ consisting of $N$ tokens. Let $E$ denote a set of pre-defined entity types, and $R$ denote a set of pre-defined relation types. Let $ {S}=\left\{s_i\right\}^{M}_{i=1}$ be all the possible continuous spans in sentence $X$. We need to predict an entity type $e \in E$ for each span $ {s_i} \in {S} $ in the sentence X. For spans of each predicted entity pair $ {s_i},{s_j} \in {S} $, we need to predict relations $ {r} \in {R} $. The output of the this task is a set of entity-relation triplets $Y=\{(({s_i},{e_i}),({s_j},{e_j}),{r}) : {s_i},{s_j} \in {S}, {e_i},{e_j} \in {E}, {r} \in {R}\}$.

\section{Model \label{sec:model}}

CARE consists of three modules:
encoder module (Sec.~\ref{ssec:encoding}), 
co-attention module  (Sec.~\ref{ssec:coattention}),
classification module (Sec.~\ref{ssec:classification}), as is shown in Figure~\ref{fig:model}.

\subsection{Encoder Module\label{ssec:encoding}}
Given a sentence $X$, the encoding module extracts contextual information from sentences. We use BERT \citep{devlin2019bert} as the pretrained encoder to produce contextual embeddings. 

\begin{equation}
    H = BERT(X)
\end{equation}

\subsection{Co-Attention Module\label{ssec:coattention}}

Typically, distinct feature and representation requisites emerge for NER and RE. The NER task may emphasize lexical-level features, whereas the RE task might prioritize contextual indications between entities. Relying solely on shared representations for these two subtasks can potentially lead to feature confusion. 

Firstly, we adopt parallel encoding, using two distinct multi-layer perceptrons (MLPs) to acquire task-specific representations for individual subtasks, thereby avoiding feature confusion. 
\begin{equation}
\begin{split}
    H_{NER} = MLP(H) \\
    H_{RE} = MLP(H)
\end{split}
\end{equation}
where $H_{NER}$ represents NER-specified representations, and $H_{RE}$ represents RE-specified representations. 

To capture spatial semantic relationships among words, we introduce a matrix $D$ for relative distance embeddings. Subsequently, we combine the task-specified representations with the relative distance embedding $D$, creating the input for the 3$\times$3 convolutional layer. Inspired by \citet{yan2021partition}, we introduce the shared representation, which can convey valuable inter-task information to both subtasks. The feature map obtained from the 3$\times$3 convolutional layer serves as the shared feature for both subtasks.

\begin{equation}
    Q = [H_{NER};H_{RE};D]
\end{equation}
\begin{equation}
    H_{share} = Conv2d(Q)
\end{equation}
For $H_{share}$, we use feed forward neural network (FFNN) and softmax activation function to calculate the co-attention score.
\begin{equation}
    A = FFNN(H_{share})
\end{equation}
\begin{equation}
    \alpha = Softmax(A)
\end{equation}
\begin{equation}
    \beta = Softmax(A^T)
\end{equation}
where $\alpha$ represents entity-to-relation attention, and $\beta$ represents relation-to-entity attention.

Next, we employ co-attention to enable interaction between task-specific representations. The results of the interaction are then integrated into the initial representations via skip connections, which are calculated as follows: 
\begin{equation}
\begin{split}
    g_{i}^{e} = h_{i}^{e} + \sum_{j=1}^n\alpha_{ij}h_{j}^{r} \\
    g_{i}^{r} = h_{i}^{r} + \sum_{j=1}^n\beta_{ij}h_{j}^{e}
\end{split}
\end{equation}
where $ h_{i}^{e}, h_{j}^{e} \in H_{NER} $, $ h_{i}^{r}, h_{j}^{r} \in H_{RE} $

By means of co-attention module, the task-specified representations $H_{NER}$ and $H_{RE}$ engage in a mutual interaction, thereby facilitating mutual enhancement between these two subtasks.

\subsection{Classification Module\label{ssec:classification}}
Following \citet{yan2021partition}, we formulate both NER and RE as table filling problems. This involves employing two-dimensional tables to represent entity labels and relation labels respectively. Note that NER can identify the entity span, the task of RE can be similarly regarded as a table filling problem, where only the head word of the entity needs to be classified. We concatenate the task-specific representations with the shared representations, which serve as the input for the NER or RE classification module.
\begin{equation}
\begin{split}
    u_{ij}^{e} = [g_{i}^{e};g_{j}^{e};h_{ij}^{s}] \\
    u_{ij}^{r} = [g_{i}^{r};g_{j}^{r};h_{ij}^{s}]
\end{split}
\end{equation}
where $h_{ij}^{s} \in H_{share}$.

The conditional probability $P(e|x_i,x_j)$ for the word pair $(x_i,x_j)$ predicting the entity type as $e$, and $P(r|x_i,x_j)$ for the same word pair predicting the relation type as $r$ are calculated as follows:
\begin{equation}
\begin{split}
    P(e|x_i,x_j) = sigmoid(W_b \sigma(W_a u_{ij}^{e} + b_a) + b_b) \\
    P(r|x_i,x_j) = sigmoid(W_d \sigma(W_c u_{ij}^{r} + b_c) + b_d)
\end{split}
\end{equation}
where $W$ and $b$ are learnable parameters, $\sigma(\cdot)$ is activation function.

Finally, we regard both NER and RE as multi-label classification problems, and use binary cross-entropy as the loss of classification.
\begin{equation}
\begin{split}
    \mathcal{L}_{NER} = -\sum_{i \leq j}\sum_{e \in E}\mathbb 
    I\left(e=1\right)logP(e|x_i,x_j)\\+ \mathbb I\left(e=0\right)log(1-P(e|x_i,x_j)) \\
    \mathcal{L}_{RE} = -\sum_{i \neq j}\sum_{r \in R}\mathbb I\left(r=1\right)logP(r|x_i,x_j)\\+\mathbb I\left(r=0\right)log(1-P(r|x_i,x_j))
\end{split}
\end{equation}
where $\mathbb I\left(\cdot\right)$ denotes indication function.

The model is trained by the sum of $\mathcal{L}_{NER}$ and $\mathcal{L}_{RE}$:
\begin{equation}
    \mathcal{L} = \mathcal{L}_{NER}+ \mathcal{L}_{RE}
\end{equation}

\section{Experiments\label{sec:experiments}}

\begin{table*}[]
\centering
\begin{tabular}{lllll}
\hline
\textbf{Dataset}                 & \textbf{Method}               & \textbf{PLM}  & \textbf{NER}  & \textbf{RE}   \\ \hline
\multirow{7}{*}{NYT}    & CopyRL \citep{zeng2019learning}               & -    & -    & 72.1 \\
                        & CasRel \citep{wei2020novel}               & B    & 93.5 & 89.6 \\
                        & TpLinker \citep{wang2020tplinker}             & B    & -    & 91.9 \\
                        & StereoRel \citep{tian2021stereorel}            & B    & -    & 92.2 \\
                        & PFN \citep{yan2021partition}                  & B    & \textbf{95.8} & 92.4 \\
                        & EmRel \citep{xu2022emrel}                & B    & -    & 92.1 \\
                        & CARE (Ours)           & B    & 95.7 & \textbf{92.6} \\ \hline
\multirow{7}{*}{WebNLG} & CopyRL \citep{zeng2019learning}               & -    & -    & 61.6 \\
                        & CasRel \citep{wei2020novel}               & B    & 95.5 & 91.8 \\
                        & TpLinker \citep{wang2020tplinker}             & B    & -    & 91.9 \\
                        & StereoRel \citep{tian2021stereorel}            & B    & -    & 92.1 \\
                        & PFN \citep{yan2021partition}                  & B    & 98.0 & 93.6 \\
                        & EmRel \citep{xu2022emrel}                & B    & -    & 92.9 \\
                        & CARE (Ours)           & B    & \textbf{98.1} & \textbf{93.9} \\ \hline
\multirow{6}{*}{SciERC} & SPE \citep{wang2020pre}                  & SciB & 68.0 & 34.6 \\
                        & UNIRE \citep{wang2021unire}                & SciB & 68.4 & 36.9 \\
                        & PURE:single-sentence \citep{zhong2021frustratingly} & SciB & 66.6 & 35.6 \\
                        & PURE:cross-sentence \citep{zhong2021frustratingly}  & SciB & 68.9 & 36.8 \\
                        & PFN \citep{yan2021partition}                  & SciB & 66.8 & 38.4 \\
                        & CARE (Ours)           & SciB & \textbf{69.9} & \textbf{40.9} \\ \hline
\end{tabular}
\caption{Performance on NYT, WebNLG and SciERC dataset. (PLM=pretrained language model, B=bert-base, SciB=scibert-base). The NER results of CasRel are reported from \citet{yan2021partition}. }
\label{table:mainresults}
\end{table*}

\subsection{Setup}
\paragraph{Datasets.}  We evaluate the performance of our model on three entity-relation extraction datasets NYT \citep{riedel2010modeling}, WebNLG \citep{gardent2017creating} and SciERC \citep{luan2018multi}. The NYT dataset is a large-scale dataset for entity-relation extraction, which is collected from news articles from The New York Times and annotated by distant supervision, including 24 relation types. The WebNLG dataset was originally created for natural language generation tasks and was applied as a relation extraction dataset by \citet{zeng2018extracting}, including 246 relation types. The SciERC dataset is collected from the abstracts of 500 artificial intelligence papers and used to build scientific knowledge graphs, including 6 entity types and 7 relation types. For all the datasets, we use the same split as the previous work \citep{yan2021partition}.
\paragraph{Evaluation Metrics.} Following the previous work \citep{yan2021partition}, we evaluate the model using partial matching on the NYT and WebNLG datasets, where only the last word of an entity is annotated in the dataset. On the SciERC dataset, models are evaluated using strict matching, where the full span of an annotated entity is used. For evaluation metrics, we report Micro-F1 scores for NER and RE respectively. For the evaluation criteria for NER, an entity is considered correct only if its type and boundary are correct. For the evaluation criteria for RE, the predicted entity-relation triples are correct only when the entity types, entity boundaries and relation types are both correct. We run three times with different random seeds on each dataset and take the average as the result.
\paragraph{Implementation details.} To ensure fair comparison with existing methods, we employ bert-base-cased \citep{devlin2019bert} for the NYT and WebNLG datasets, and scibert-scivocab-cased \citep{beltagy2019scibert} for the SciERC dataset.

\subsection{Main Results}
Table~\ref{table:mainresults} shows a comprehensive comparison between our model and existing methods. In both NER and RE tasks, when compared against the baseline model CasRel \citep{wei2020novel}, CARE shows notable performance improvements. On the NYT dataset, CARE shows improvements of 2.2\% and 3.0\% for NER and RE respectively. On the WebNLG dataset, CARE outperforms CasRel with gains of 2.6\% for NER and 2.1\% for RE. On the SciERC dataset, CARE also achieve superior performance, outperforming PFN \citep{yan2021partition} by 3.1\% for NER and 2.5\% for RE. What sets CARE apart from the majority of models is its capacity to capture the bidirectional interaction between entities and relations, achieved through the co-attention module. The aforementioned enhancements suggest that utilizing information from one subtask to augment the other yields significant benefits.

\subsection{Ablation Study}

We investigate the efficacy of various components within the model. These components include the relative distance embedding, co-attention mechanism, shared representation, 3${\times}$3 convolutional layers. The ablation results on the SciERC dataset are shown in Table~\ref{table:ablation}. We find that: (1) Without the relative distance embeddings, there is a 0.6$\%$ decrease in NER and a 0.5$\%$ decrease in RE, which verified their useful in capturing the spatial relationships between words. (2) In the case of removing the shared representation from the classification layer, there is a decline of 0.5$\%$ in NER and 0.9$\%$ in RE, indicating the presence of interdependencies and correlations between subtasks. (3) When we replace the 3$\times$3 convolutional layers with 1$\times$1 convolutional layers, the performance of both NER and RE dropped. The 3$\times$3 convolutional layers, with larger receptive field, are better suited to capture localized contextual information. (4) When we remove the co-attention module, a noticeable performance decline is observed. Specifically, there is a reduction of 1.2$\%$ in NER and 1.3$\%$ in RE, indicating that the co-attention module can effectively capture the two-way interaction between NER task and RE task. 

Moreover, we investigate the influence of the co-attention module depth, as shown in Table~\ref{table:ablation2}. Notably, the best performance is achieved when the co-attention module comprises 3 layers. As the number of layers increases to 4, a reduction in performance becomes evident. It could potentially be attributed to the increased risk of overfitting as the number of layers grows, causing the loss of certain vital low-level features.

\begin{table}[]
\footnotesize
\centering
\begin{tabular}{lllll}
\hline
\textbf{Setting}                   & \textbf{NER}  & \textbf{$\Delta$}   & \textbf{RE}   & \textbf{$\Delta$}   \\ \hline
Default                   & 69.9 & -   & 40.9 & -   \\
- Distance embedding    & 69.3 & 0.6 & 40.4 & 0.5 \\
- Shared representation & 69.4 & 0.5 & 40.0 & 0.9 \\
- 3$\times$3 convolution       & 69.1 & 0.8 & 40.2 & 0.7 \\ 
- Co-attention          & 68.7 & 1.2 & 39.6 & 1.3 \\ \hline
\end{tabular}
\caption{Ablation study on the SciERC dataset.}
\label{table:ablation}
\end{table}

\begin{table}[]
\footnotesize
\centering
\begin{tabular}{lllll}
\hline
\textbf{Layers}                   & \textbf{NER}   & \textbf{RE}    \\ \hline
N=1                   & 69.5   & 40.3  \\
N=2    & 69.6 & 40.6  \\
N=3          & \textbf{69.9}  & \textbf{40.9} \\
N=4 & 68.2  & 40.5 \\ \hline
\end{tabular}
\caption{The impact of co-attention module depths on the SciERC dataset.}
\label{table:ablation2}
\end{table}

\begin{figure*}
    \centering
    \includegraphics[width=1.0\linewidth]{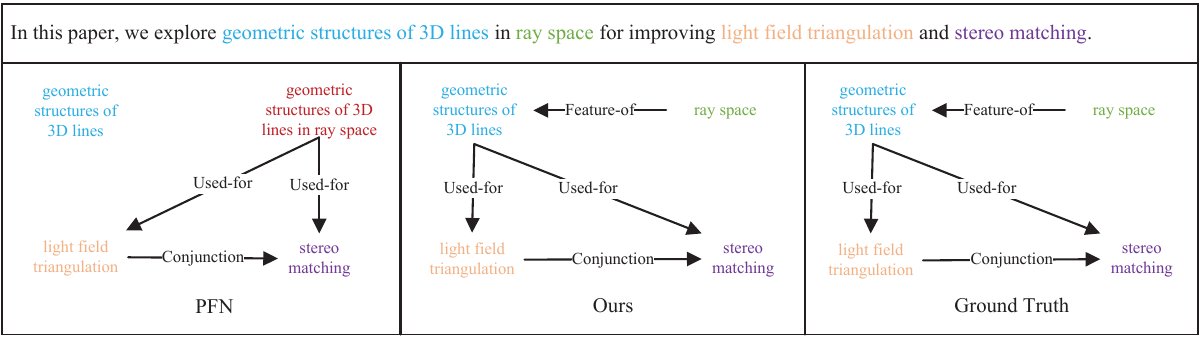}
    \caption{Case study on the SciERC test set. Entities are distinguished by being assigned various colors.}
    \label{fig:casestudy}
\end{figure*}

\subsection{Case Study}
We present a case study on the SciERC test set and make a comparison with the superior-performing model PFN \citep{yan2021partition} to demonstrate the advantages of CARE within specific scenarios. As shown in Figure~\ref{fig:casestudy}, obviously, the PFN fails to recognize the entity "ray space" and the relation triplet (\textquotedbl ray space\textquotedbl, \textquotedbl Feature-of\textquotedbl, \textquotedbl geometric structures of 3D lines\textquotedbl), and the entity \textquotedbl ray space\textquotedbl is wrongly associated with another entity \textquotedbl geometric structures of 3D lines\textquotedbl. In contrast, CARE successfully recognizes all entities and relations within the sentence, demonstrating a strong ability to harness the interaction between the two subtasks. Compared with the PFN, the utilization of relative distance embedding and 3$\times$3 convolutions in our model enhances the capacity to capture interactions at the word-pair level, leading to improved accuracy in identifying entity boundaries.

\section{Related Work}

Most joint models can be classified into the following four categories: labeling-based, generation-based, span-based and table filling.

\paragraph{Labeling-based.} \citet{bekoulis2018joint} adopt the BIO (Beginning, Inside, Outside) encoding scheme for NER, and regard RE as a multi-head selection problem. \citet{wei2020novel} propose a cascaded binary labeling framework, wherein the head entity is initially extracted, followed by the extraction of the relation and its associated tail entity.

\paragraph{Generation-based.} \citet{zeng2018extracting} propose a Seq2Seq relation extraction framework based on a copy mechanism. \citet{zeng2019learning} apply reinforcement learning to a Seq2Seq model, considering the ordering of relation extraction. \citet{cabot2021rebel} view relation extraction as a Seq2Seq task, utilizing the pretrained model BART \citep{lewis2020bart} to generate relation triplets from the original sentence inputs.

\paragraph{Span-based.} \citet{luan2019general} regard NER as a span classification problem and construct dynamic span graphs to produce span representations. \citet{eberts2020span} use max-pooling to produce span-level representations. \citet{wang2020pre} integrate span-level information into pretrained model for entity-relation extraction.

\paragraph{Table filling.} \citet{wang2020tplinker} formulates joint extraction as a word-pair linking problem and introduced a handshake labeling scheme that aligns boundary words of entity pairs under each relation type. \citet{wang2020two} designed two different encoders to capture table features and sequence features respectively. \citet{wang2021unire} unify the label spaces of the two subtasks in a table. \citet{yan2021partition} proposed a partition filtering network to model bidirectional interactions between subtasks. 

Compared to PFN \citep{yan2021partition} and other approaches to entity-relation extraction, our proposed method features many different designs, each with unique motivations. First, the ways of learning task-specified features are different. We employ a parallel encoding approach, using two task-specific MLPs to extract task-specific features, whereas PFN only utilizes a single RNN-like encoder introducing entity and relation gates to filter out task-specific features. Specifically, the parallel encoding strategy enables direct learning of task-specific features for each subtask without extra filtering operations, thereby avoiding feature overlap. Second, the ways of modeling two-way interaction are different. In traditional parallel encoding approaches, task-specific features are independently generated using shared input. In order to model the interaction between the two subtasks based on parallel encoding, we introduces a co-attention mechanism to selectively fuse all possible information from one subtask to enhance the other, thereby promoting mutual reinforcement.

\section{Conclusion}
In this paper, we propose CARE, a co-attention network designed for joint entity and relation extraction. We adopt parallel encoding to separately learn task-specific representations for NER and RE, preventing feature confusion between the two subtasks. Building upon parallel encoding, we introduce a co-attention mechanism to capture two-way interaction between NER and RE, effectively leveraging the information from one subtask to enhance the other. We evaluate our model through extensive experiments on three widely-used benchmark datasets (NYT, WebNLG and SciERC). The experimental results demonstrate that CARE achieves significant improvements over existing baseline methods.

\section{Acknowledgements}

We thank all the anonymous reviewers for their helpful comments and feedback. This work was supported in part by the National Key Research and Development Program of China under Grant 2022YFF0902701, in part by the National Natural Science Foundation of China under Grants U21A20468, 62372058, U22A201339, 62162060, in part by the Fundamental Research Funds for the Central Universities under Grant 2020XD-A07-1.

\section{Bibliographical References}\label{sec:reference}
\bibliographystyle{lrec-coling2024-natbib}
\bibliography{lrec-coling2024-example}

\end{document}